%% file: main.tex
\documentclass[10pt,twocolumn,letterpaper]{article}

\usepackage{cvpr}              

\input{preamble}

\definecolor{cvprblue}{rgb}{0.21,0.49,0.74}
\usepackage[pagebackref,breaklinks,colorlinks,citecolor=cvprblue]{hyperref}
\usepackage{graphicx}
\usepackage{amsmath}
\usepackage{amssymb}
\usepackage{booktabs}
\usepackage{multirow}
\usepackage{utfsym}
\usepackage{color}
\usepackage{mathtools, nccmath}
\usepackage[misc]{ifsym}
\usepackage[hypcap=false]{caption}
\usepackage{footnote}
\usepackage{footmisc}

\title{VividTalk: One-Shot Audio-Driven Talking Head Generation \\ Based on 3D Hybrid Prior
}

\author{
Xusen Sun$^{1,*}$ \and
Longhao Zhang$^3$ \and
Hao Zhu${\textsuperscript{1, \Letter}}$ \and 
Peng Zhang${\textsuperscript{2, \Letter}}$ \and
Bang Zhang$^2$ \and 
Xinya Ji$^1$ \and
Kangneng Zhou$^4$ \and
Daiheng Gao$^2$ \and
Liefeng Bo$^2$ \and
Xun Cao$^1$  \\
$^{1}$Nanjing University \qquad $^{2}$Alibaba Group \qquad $^{3}$ByteDance \qquad $^{4}$Nankai University \\
\\
\href{https://humanaigc.github.io/vivid-talk/}{https://humanaigc.github.io/vivid-talk/}
}

\begin{document}

\twocolumn[{%
\renewcommand\twocolumn[1][]{#1}%

\maketitle
\vspace{-0.35in}
\begin{center}
    \includegraphics[width=1.0\linewidth]{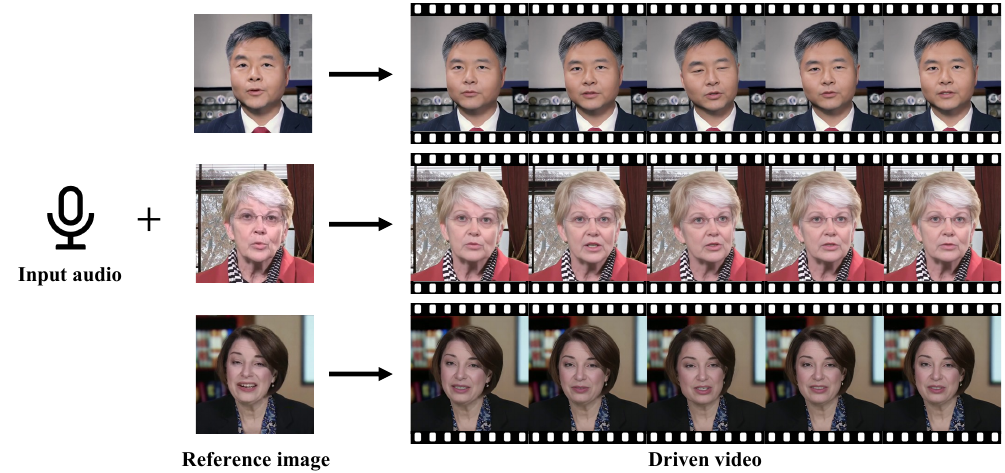}
    \vspace{-0.25in}
    \captionof{figure}{We proposed VividTalk, a generic talking head generation framework. Our method can generate high-visual quality talking head videos with expressive facial expressions, various head poses, and lip-sync enhanced by a large margin. 
    }
    \label{fig:title}
\end{center}
}]

\let\thefootnote\relax\footnotetext{$^*$ Work done as an intern at Alibaba Group.}

\input{sec/0_abstract}    
\input{sec/1_intro}
\input{sec/2_related}
\input{sec/3_method}
\input{sec/4_exp}
\input{sec/5_con}


{
    \small
    \bibliographystyle{ieeenat_fullname}
    \bibliography{main}
}

\end{document}

%% file: preamble.tex
\usepackage[dvipsnames]{xcolor}


%% file: sec/0_abstract.tex
\begin{abstract}
Audio-driven talking head generation has drawn much attention in recent years, and many efforts have been made in lip-sync, expressive facial expressions, natural head pose generation, and high video quality. However, no model has yet led or tied on all these metrics due to the one-to-many mapping between audio and motion. In this paper, we propose VividTalk, a two-stage generic framework that supports generating high-visual quality talking head videos with all the above properties. Specifically, in the first stage, we map the audio to mesh by learning two motions, including non-rigid expression motion and rigid head motion. For expression motion, both blendshape and vertex are adopted as the intermediate representation to maximize the representation ability of the model. For natural head motion, a novel learnable head pose codebook with a two-phase training mechanism is proposed. In the second stage, we proposed a dual branch motion-vae and a generator to transform the meshes into dense motion and synthesize high-quality video frame-by-frame. Extensive experiments show that the proposed VividTalk can generate high-visual quality talking head videos with lip-sync and realistic enhanced by a large margin, and outperforms previous state-of-the-art works in objective and subjective comparisons.

\end{abstract}

%% file: sec/1_intro.tex
\section{Introduction}

One-shot audio-driven talking head generation aims to drive an arbitrary facial image with audio as input signal and has extensive application scenarios, such as virtual avatars \cite{thies2020neural, song2018talking, gu2020flnet}, visual dubbing \cite{kr2019towards, prajwal2020lip, xie2021towards}, and video conferences \cite{chen2020talking, wang2021one, zakharov2019few, zhang2020davd, zhou2020makelttalk}. As a consequence, it has attracted widespread attention and inspired many researchers to work in this field. 

The facial motion of a talking head mainly comes from two folds: non-rigid facial expression components and rigid head components. To maximize the photo-realism of the generated videos, both components need to be taken into consideration. For facial expression motion, most existing approaches adopt a multi-stage framework to map the audio feature to an intermediate representation, \textit{e.g.}, facial landmarks \cite{zhou2020makelttalk, zhong2023identity}, and 3DMM coefficients \cite{zhang2023sadtalker, zhang2021flow}. However, the facial landmarks are too sparse to model the expressive facial expression in detail. By contrast, the 3D face morphable model \cite{blanz2023morphable} (3DMM) has been proven to have the ability to represent the face with various expressions. Whereas, we observed that the distribution of blendshapes on the same expression varies considerably, which exacerbates the one-to-many mapping problem between audio and facial motion and leads to a lack of fine-grained motion. For rigid head motion, it is harder to model because of the weak relationship with audio. Some works \cite{prajwal2020lip, zhou2021pose, wang2023seeing} utilize a video to provide the head pose or to keep the head still when speaking. Another line of methods \cite{zhou2020makelttalk, zhang2023sadtalker, zhang2021flow} present to learn head poses from audio directly, but generate noncontinuous and unnatural results. Up to now, how to generate reasonable head poses from audio is still a challenging problem to be solved.

To address the above problems, we proposed VividTalk, a generic one-shot audio-driven talking head generation framework. Our method only takes a single reference facial image and an audio sequence as inputs, then generates a high-quality talking head video with expressive facial expressions and various head poses. Specifically, the proposed model is a two-stage framework consisting of Audio-To-Mesh Generation and Mesh-To-Video Generation. In the first stage, considering the one-to-many mapping between facial motion and blendshape distribution, we utilize both blendshape and 3D vertex as the intermediate representation, in which blendshape provides a coarse motion and vertex offset describes a fine-grained lip motion. Besides, a multi-branch transformer-based network is also adopted to make full use of long-term audio context to model the relation with the intermediate representations. To learn rigid head motion from audio more reasonably, we cast this problem as a code query task in a discrete and finite space, and build a learnable head pose codebook with a reconstruction and mapping mechanism. After that, both motions learned are applied to reference identity, resulting in driven meshes. In the second stage, based on the driven meshes and reference image, we render the projection texture of both the inner face and outer face, such as the torso, to model the motion comprehensively. Then a novel dual branch motion-vae is designed to model the dense motion, which is fed as input to a generator to synthesize the final video in a frame-by-frame manner. 

Extensive experiments show that our proposed VividTalk can generate lip-sync talking head videos with expressive facial expressions and natural head poses. As shown in Figure~\ref{fig:title} and Table~\ref{tab: quantitative_comparison}, both visual results and quantitative analysis demonstrate the superiority of our method in both generated quality and model generalization. To summarize, the
main contributions of our work are as follows:

\begin{itemize}
    \item We present to map the long-term audio context to both blendshape and vertex to maximize the representation capability of the model, and an elaborate multi-branch generator is designed to model global and local facial motions individually. 
    \item A novel learnable head pose codebook with a two-phase training mechanism is proposed to model the rigid head motion more reasonably. 
    \item Experiments demonstrate that our proposed VividTalk is superior to the state-of-the-art methods, supporting high-quality talking head video generation and can be generalized across various subjects. 
\end{itemize}

%% file: sec/2_related.tex
\section{Related works}

\noindent \textbf{Audio-driven talking head generation. } Audio-driven talking head generation aims to drive a facial image according to the audio signal. Early works ~\cite{chen2018lip, vougioukas2020realistic, chung2017you} tried to generate videos in an end-to-end manner. Recently, some works adopted a multi-stage framework to map audio to an intermediate representation, such as 3DMM coefficients ~\cite{zhang2023sadtalker, zhang2021flow, ren2021pirenderer}, and facial landmarks ~\cite{gururani2022spacex, zhou2020makelttalk, zhong2023identity}, to model the motion better. ~\cite{zhang2023sadtalker} first generates the 3DMM coefficients from audio, and then the generated 3DMM coefficients are mapped to the unsupervised 3D keypoints to modulate the face render to synthesize videos. ~\cite{ren2021pirenderer} proposes to control the facial motions with 3DMM coefficients and generates final image in an coarse-to-fine strategy. Its framework can be easily extended to tackle audio-driven talking head tasks by learning a mapping from audio to 3DMM coefficients. ~\cite{gururani2022spacex} uses facial landmarks and a pre-trained face render to make the generated talking head videos more controllable and high-quality. Similarly, facial landmarks are predicted by ~\cite{zhou2020makelttalk} to reflect the speaker-aware dynamics to animate both human face images and non-photorealistic cartoon images. ~\cite{zhong2023identity} only generates lip-related landmarks to inpaint the lower-half occluded facial images. Besides, multiple reference images are needed to produce realistic rendering. However, all of these methods are insufficient to generate lip-sync and realistic talking head videos because of the limitation of the intermediate representation. By contrast, our method uses both blendshape and vertex as the intermediate representation to model the coarse motion and fine-grained motion, respectively. 

\noindent \textbf{Video-driven talking head generation.} Video-driven talking head generation focuses on transferring the motion of the source actor to the target subject, which is also known as face reenactment. The approaches generally fall into two categories: subject-specific and subject-agnostic. Subject-specific methods \cite{suwajanakorn2017synthesizing, thies2015real, thies2016face2face} can produce high-quality videos but can not be extended to new subjects, which limits their application. Recently, some subject-agnostic works \cite{siarohin2019first, wang2021one, hong2022depth, wu2022anifacegan, ren2021pirenderer, tripathy2021facegan} have tried to address this problem and achieved tremendous success. For example, ~\cite{siarohin2019first} disentangles the appearance and motion self-supervised, and learn keypoints along with their local affine transformations to animate the source image. ~\cite{hong2022depth} proposes to recover the explicit dense 3D geometry from videos and utilizes the learned depth information to improve the performance of generated talking head videos. Compared to the above methods, our task is more challenging because we need to drive the image with audio as input without any motion prior knowledge.

%% file: sec/3_method.tex
\section{Method}

\begin{figure*}[th]
    \centering
    \includegraphics[width=1.0\linewidth]{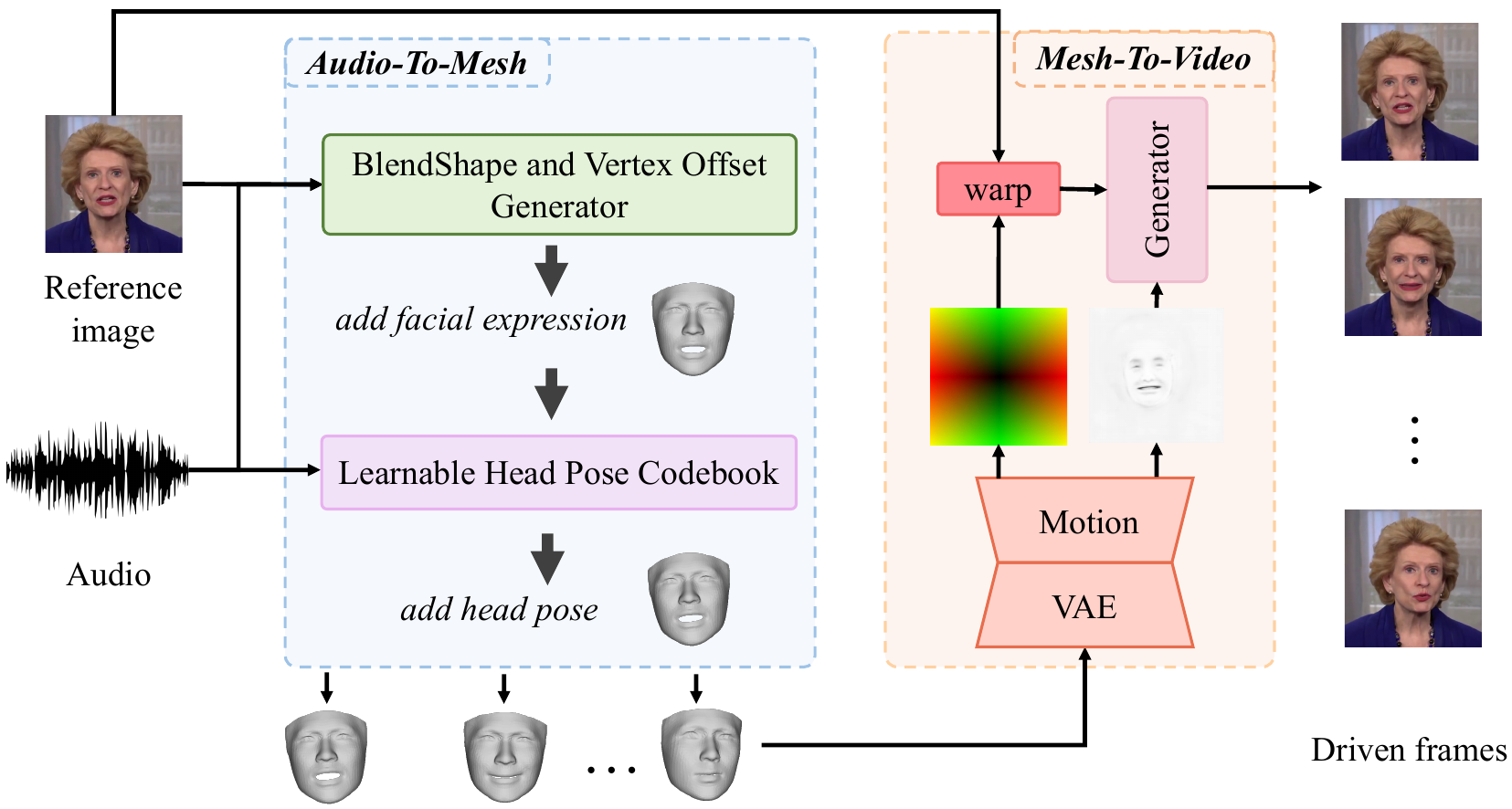}
    \vspace{-0.2in}
    \caption{Overview of the proposed VividTalk.  Our framework is constituted by two cascaded stages. The Audio-To-Mesh stage maps the audio to non-rigid facial expression motion and rigid head pose, respectively, which results in the driven meshes. The Mesh-To-Video stage transforms the driven meshes into 2D dense motion and synthesizes high-visual quality and realistic talking head videos. 
    }   
    \vspace{-0.1in}
    \label{fig: framework}
\end{figure*}

Our method can generate talking head videos with diverse facial expressions and natural head poses given an audio sequence and a reference facial image as input. As shown in Figure \ref{fig: framework}, our framework is composed of two cascaded stages, named Audio-To-Mesh Generation and Mesh-To-Video Generation, respectively. 
In the following, we first briefly introduce some preliminaries of the 3D morphable model and data preprocessing in Section \ref{method_preliminaries}. Then, the design of the Audio-To-Mesh stage and Mesh-To-Video stage are described in Section \ref{method_audio_to_mesh} and Section \ref{method_mesh_to_video}, respectively. Finally, we depicted the training strategy of the total framework in Section \ref{method_training_strategy}. 

\subsection{Preliminaries}
\label{method_preliminaries}
\noindent \textbf{3D Morphable Model.} Our method uses 3D-based (blendshape and vertex) instead of 2D-based information as the intermediate representation for talking head generation. In 3DMM \cite{blanz2023morphable}, the 3D face shape can be represented as: 

\begin{equation}
    S=\overline{S} + \alpha U_{id} + \beta U_{exp},
\end{equation}

\noindent where $\overline{S}$ is the mean shape of the face, $U_{id}$, and $U_{exp}$ are the PCA bases of identity and expression, respectively. $\alpha$ and $\beta$ are the identity and expression coefficients for generating a 3D face.   

\noindent \textbf{Data Preprocessing.} Our model only needs to be trained with an audio-visual synchronized dataset. Before training, some data preprocessing is a prerequisite. Specifically, given a talking head video, we first crop the face region and resize it into $256 \times 256$ following ~\cite{siarohin2019first}. Then the coefficients $\{\alpha \in \mathbb{R}^{150},\beta \in \mathbb{R}^{52} \}^{\times f}$ and mesh vertices sequence $M^{(3 \times n) \times f}$ are reconstructed by ~\cite{wang2022faceverse}, where $n$ is the vertex number and $f$ is the frame number. To model the head pose $P$, rotation matrix $R \in \mathbb{SO}(3)$ and translation vector $t \in \mathbb{R}^3$ are also extracted.

\subsection{Audio-To-Mesh Generation}
\label{method_audio_to_mesh}
In this section, our goal is to generate 3D-driven meshes according to the input audio sequence and a reference facial image. To be more specific, we first utilize FaceVerse\cite{wang2022faceverse} to reconstruct the reference facial image. Next, we learn both non-rigid facial expression motion and rigid head motion from the audio to drive the reconstructed mesh. To this end, a multi-branch BlendShape and Vertex Offset Generator and a Learnable Head Pose Codebook are proposed. 

\noindent \textbf{BlendShape and Vertex Offset Generator.} Learning a generic model to generate accurate mouth movements and expressive facial expressions with person-specific style is challenging in two aspects: 1) The first challenge is the \textit{audio-motion correlation} problem. As audio signal correlates best with mouth movements, it is difficult to model non-mouth motion from audio. 2) The mapping from audio to facial expression motions naturally has one-to-many properties, which means that the same audio input may have more than one correct motion pattern, leading to a \textit{mean face} phenomenon with no personal characteristics. To solve the \textit{audio-motion correlation} problem, we use both blendshape and vertex offset as the intermediate representation, for which blendshape provides a coarse facial expression motion globally and lip-related vertex offset offers a fine-grained lip motion locally. As for the \textit{mean face} problem, we proposed a multi-branch transformer-based generator to model each part's motion individually and inject the subject-specific style to maintain personal features. 

\begin{figure}[th]
    \centering
    \includegraphics[width=1.0\linewidth]{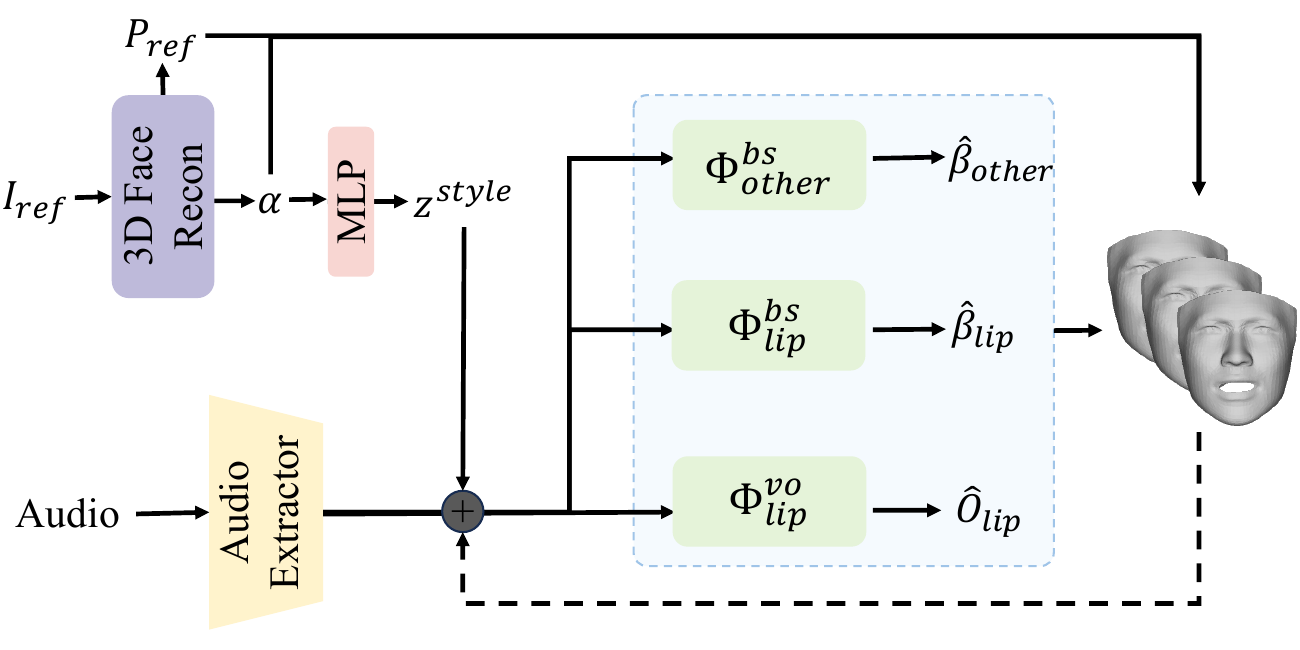}
    \vspace{-0.2in}
    \caption{The structure of the proposed BlendShape and Vertex Offset Generator. The blendshape $\{ \hat{\beta}_{lip}$, $\hat{\beta}_{other} \}$ provide the coarse facial expression motion with personal style, and the lip-related vertex offset $\hat{O}_{lip}$ supplement the lip motion at a fine-grained level. 
    }   
    \vspace{-0.1in}
    \label{fig: blendshape_vertices_offset}
\end{figure}

Specifically, we utilize a pre-trained audio extractor \cite{baevski2020wav2vec} to extract the contextualized speech representation $A=(a_{1}, a_{2}, ..., a_{f})$ from the input audio sequence. To represent the person-specific style characteristic, a pre-trained 3D face reconstruction model \cite{wang2022faceverse} is used to extract the identity information $\alpha$ from the reference image $I_{ref}$, which will be encoded as a style embedding $z^{style}$. Then the audio feature $A$ and the personal style embedding $z^{style}$ are added and fed into a multi-branch transformer-based architecture with two branches to generate blendshape that models facial expression motion at a coarse level, and the third branch to generate lip-related vertex offset as supplementation of lip motion at a fine-grained level. Note that to model the temporal dependencies better, the learned past motions will be taken as the input of the network when predicting the current motion, which can be formulated as

\begin{equation}
    \hat{\beta}^{f}_{i}=\Phi_{i}^{bs}(\hat{\beta}^{1...f-1}_{i}, A, z^{style}), \quad i \in \{lip, other\},
\end{equation}

\begin{equation}
    \hat{O}^{f}_{lip}=\Phi_{lip}^{vo}(\hat{O}^{1...f-1}_{lip}, A, z^{style}),
\end{equation}

\noindent where $\hat{\beta}_{lip}^{f}$, $\hat{\beta}_{other}^{f}$ are the lip-related blendshape and the other blendshape at frame $f$, respectively. $\hat{O}^{f}_{lip}$ is the lip-related vertex offset at frame $f$. And $\Phi$ is the corresponding network of each branch. Once the training is finished, the driven meshes with non-rigid facial expression motion can be obtained by

\begin{equation}
    \hat{M}_{nr} = (\overline{S} + \alpha U_{id} + (\hat{\beta}_{lip} , \hat{\beta}_{other}) U_{exp} + \hat{O}_{lip}) \otimes P_{ref},
\end{equation}

\noindent where $P_{ref}$ is the pose of reference facial image and $\otimes$ represents the affine transformation caused by $P_{ref}$. 

\begin{figure}[t]
    \centering
    \includegraphics[width=1.0\linewidth]{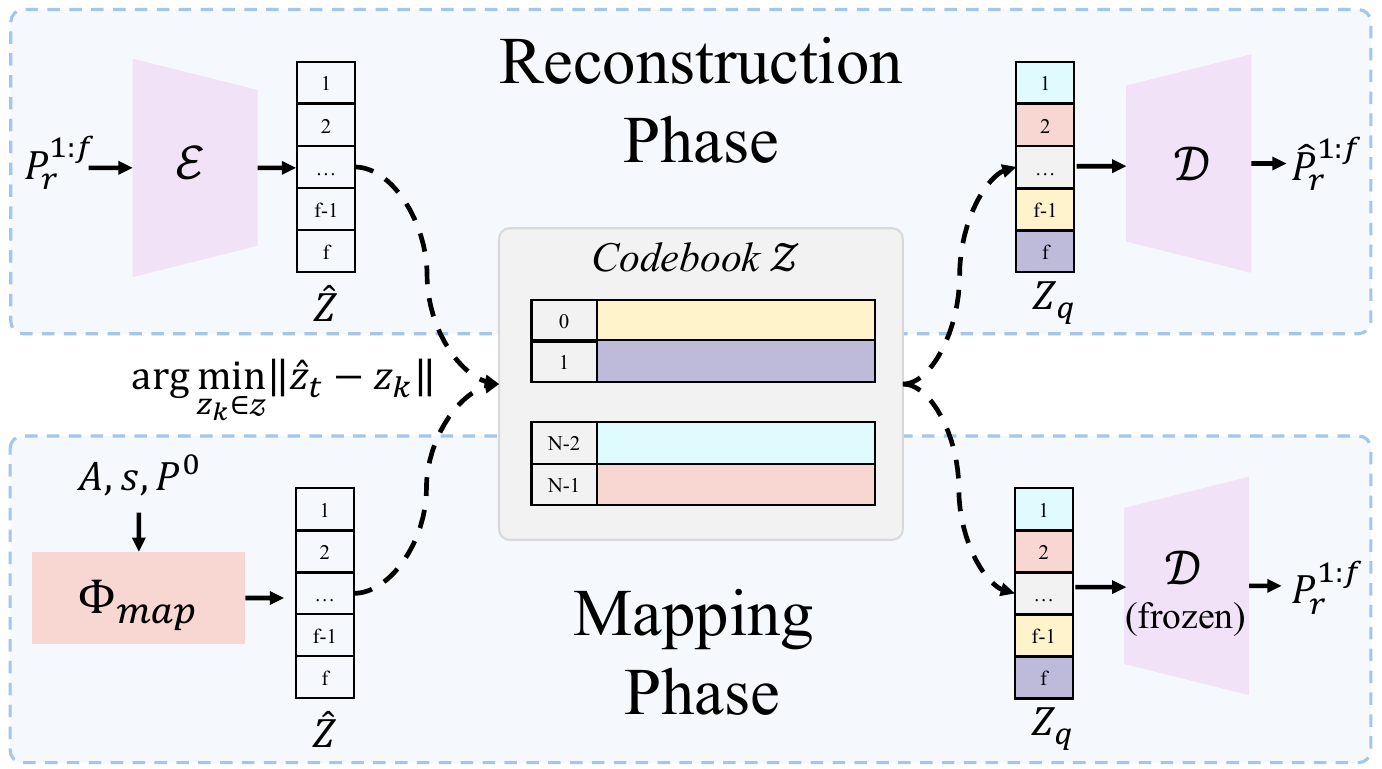}
    \vspace{-0.2in}
    \caption{The two-phase training mechanism of the Learnable Head Pose Codebook. Note that we train the two phases separately, the decoder $\mathcal{D}$ and codebook $\mathcal{Z}$ are frozen during the training of the mapping phase. 
    }   
    \vspace{-0.1in}
    \label{fig: learnable_head_pose}
\end{figure}

\noindent \textbf{Learnable Head Pose Codebook.} The head pose is another important factor that influences the realism of talking head videos. However, it is not easy to learn it from audio directly because of the weak relationship between them, which will lead to unreasonable and discontinuous results. Inspired by ~\cite{van2017neural} which utilized a discrete codebook as a prior to guarantee high-fidelity generation even with a degraded input. We propose to cast this problem as a code query task in a discrete and finite head pose space and a two-phase training mechanism is carefully designed, with the first phase building an abundant head pose codebook and the second phase mapping the input audio to the codebook to generate the final results, as shown in Figure~\ref{fig: learnable_head_pose}.

In the reconstruction phase, the task is to build a context-rich head pose codebook $\mathcal{Z}=\{z_{k}\}_{k=1}^{K}$ and a decoder $\mathcal{D}$ with the ability to decode realistic head pose sequence $P^{1:f} \in \mathbb{R}^{6 \times f}$ from $\mathcal{Z}$. We adopt a VQ-VAE which constitutes an encoder $\mathcal{E}$, a decoder $\mathcal{D}$, and a codebook $\mathcal{Z}$ as the backbone. Firstly, the relative head pose $P_{r}^{1:f}=P^{1:f}-P^{0}$ is calculated and encoded as a latent code $\hat{Z}=\mathcal{E}(P_{r}^{1:f})$. Then we obtain $Z_{q}$ using an element-wise quantization function $\mathbf{q}(\cdot)$ to map each item $\hat{z}$ in $\hat{Z}$ to its closest codebook entry $z_{k}$:

\begin{equation}
    Z_{q}=\mathbf{q}(\hat{z})= \underset{z_{k} \in \mathcal{Z}}{{\arg\min}} \left\| \hat{z} - z_{k} \right\| \label{quantization function}.
\end{equation}

Finally, based on the $Z_{q}$, the reconstructed relative head pose $\hat{P}_{r}^{1:f}$ is given by the decoder $\mathcal{D}$ as follows: 

\begin{equation}
    \hat{P}_{r}^{1:f}=\mathcal{D}(Z_{q})=\mathcal{D}(\mathbf{q}(\mathcal{E}(P_{r}^{1:f}))).
\end{equation} 

In the mapping phase, we focus on building a network that can map the audio to the codebook learned in the previous phase to generate natural and successive head pose sequences. To model the temporal continuity better, a transformer-based autoregressive model $\Phi_{map}$ with self-attention and cross-modal multi-head attention mechanisms was proposed. Specifically, $\Phi_{map}$ takes an audio sequence $A$, person-specific style embedding $z^{style}$ and initial head pose $P^{0}$ as input, and output an  intermediate feature $\hat{Z}$ which will be quantized into $Z_{q}$ from codebook $\mathcal{Z}$, and then decoded by the pre-trained decoder $\mathcal{D}$:

\begin{equation}
    \hat{P}_{r}^{1:f}=\mathcal{D}(Z_{q})=\mathcal{D}(\mathbf{q}(\Phi_{map}(A, s, P^{0}))).
\end{equation}

\noindent Note that the codebook $\mathcal{Z}$ and the decoder $\mathcal{D}$ are frozen during the training of mapping phase. 

So far, both the non-rigid facial expression motion and rigid head pose have been learned. Now, we can obtain the final driven meshes $\hat{M}_{d}$ by applying the learned rigid head pose to mesh $\hat{M}_{nr}$:

\begin{equation}
    \hat{M}_{d}^{1:f} = \hat{M}_{nr}^{1:f} \otimes \hat{P}_{r}^{1:f}.
\end{equation}

\subsection{Mesh-To-Video Generation}
\label{method_mesh_to_video}
\begin{figure}[th]
    \centering
    \includegraphics[width=1.0\linewidth]{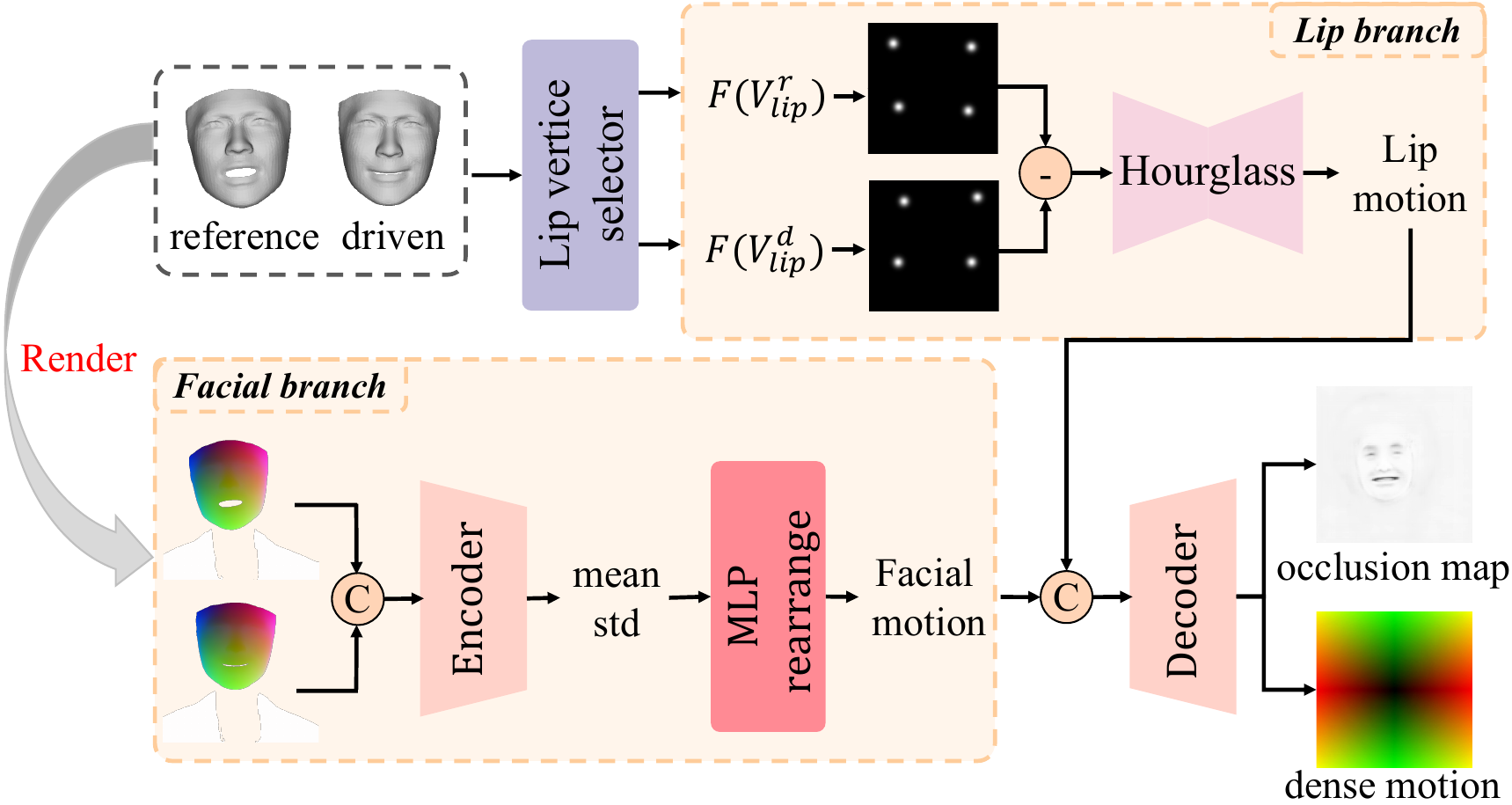}
    \vspace{-0.2in}
    \caption{The architecture of the proposed dual branch motion-vae. The below branch models the motion across images globally. The upper branch augments the lip motion based on lip-related landmarks. 
    }   
    \vspace{-0.1in}
    \label{fig: mesh_to_video}
\end{figure}

This section is devoted to transforming the driven meshes into videos. As shown in Figure ~\ref{fig: mesh_to_video}, a dual branch motion-vae is proposed to model the 2D dense motion, which will be taken as the input of the generator to synthesize the final video. Next, we will introduce this process in detail. 

Transforming 3D domain motion to 2D domain motion directly is difficult and inefficient because the network needs to seek the correspondence between two domain motions for better modeling. To decrease the learning burden of the network and achieve further performance, we conduct this transformation in the 2D domain with the help of projection texture representation. 

To render the projection texture of 3D mesh, we first normalize the mean shape of the 3D face to $0-1$ in $x, y, z$ axis to obtain a Normalized Coordinate Code $NCC$ with three channels similar to RGB, which can be seen as a new representation of the face texture:

\begin{equation}
    NCC_{i} = \frac{\overline{S}_{i} - min(\overline{S}_{i})}{max(\overline{S}_{i}) - min(\overline{S}_{i})}, \quad i \in \{x, y, z\}.
\end{equation}

\noindent Then we adopt Z-Buffer to render the projected 3D inner face texture $PT_{in}$ colored by $NCC$. However, the outer face region can not be modeled well because of the limitation of 3DMM. To model the motion across frames better, we use \cite{liu2015deep} to parse images and obtain the outer face region texture $PT_{out}$, such as the torso and background, which will be combined with $PT_{in}$ as below:

\begin{equation}
    PT = PT_{in} \cdot M + PT_{out} \cdot (1 - M)
\end{equation}

\noindent where $M$ is the mask of the inner face. 

As shown in Figure ~\ref{fig: mesh_to_video}, in the facial branch, the reference projected texture $PT_{ref}$ and driven projected texture $PT_{d}$  are concatenated and fed into an Encoder followed by an MLP, which outputs a 2D facial motion map. To further enhance lip movements and model more accurately, we also selecte lip-related landmarks and transform them into Gaussian maps, a more compact and efficient representation. Then an Hourglass network takes the substracted Gaussian map as input and outputs a 2D lip motion, which will be concatenated with the facial motion and decoded into a dense motion and an occlusion map. 

Finally, we warp the reference image based on the dense motion map predicted before and obtain the deformed image, which will be taken as the input to the generator with the occlusion map to synthesize the final video frame by frame.

\subsection{Training Strategy}
\label{method_training_strategy}
Training such a framework is not easy. Specifically, we train the Audio-To-Mesh stage and Mesh-To-Video stage separately. And the complete framework can be inferred in an end-to-end fashion. The BlendShape and Vertex Offset Generator are supervised by reconstruction loss in terms of blendshape and mesh:

\begin{equation}
    L_{bsvo} = \left\| \beta - \hat{\beta} \right\| + \left\| M - \hat{M}_{nr} \right\|.
\end{equation}

In the training of Learnable Head Pose Codebook, due to the quantization function ~\ref{quantization function} is not differentiable, we apply a straight-through gradient estimator~\cite{bengio2013estimating} that copies the gradients from the decoder to the encoder. Then the two-phase training is supervised as follows:

\begin{equation}
    \begin{aligned}
        L_{rec}= & \left\| P_{r}^{1:f} - \hat{P}_{r}^{1:f} \right\|^{2} + \left\| sg(\mathcal{E}(P_{r}^{1:f})) - z_{q} \right\|_{2}^{2} \\
                    & + \left\| sg(z_{q}) - \mathcal{E}(P_{r}^{1:f}) \right\|_{2}^{2},
    \end{aligned}
\end{equation}

\begin{equation}
    L_{map}=\left\| P_{r}^{1:f} - \hat{P}_{r}^{1:f} \right\|^{2} 
    + \left\| \hat{Z} - sg(Z_{q}) \right\|_{2}^{2} ,
\end{equation}

\noindent where $sg(\cdot)$ denotes a stop-gradient operation. 

As for the Mesh-To-Video stage, the perceptual loss $L_{perc}$ based on the pre-trained VGG-19 \cite{simonyan2014very} network is used as the main driving loss.  The feature matching loss $L_{fm}$ is also used to stabilize the training as the generator has to produce realistic results. 

%% file: sec/4_exp.tex
\section{Experiments}

\subsection{Dataset and Metrics}

\noindent \textbf{Dataset.} We train our model with the HDTF~\cite{zhang2021flow} dataset and VoxCeleb~\cite{nagrani2017voxceleb} dataset. HDTF is a high-resolution audio-visual dataset containing over 16 hours of video on 346 subjects. VoxCeleb is another larger dataset involving more than 100k videos and 1000 identities. We first filter the two datasets to remove the invalid data, \textit{e.g.}, data with out-of-sync audio and video. Then following the~\cite{siarohin2019first}, we leverage a face landmarks detector to crop the face region in the video and resize them into $256 \times 256$. Finally, the processed videos are divided into $80\%, 10\%, 10\%$, which will be used for training, validating, and testing. 

\noindent \textbf{Metrics.} To demonstrate the superiority of the proposed method, we evaluate the model with several metrics. The SyncNet score ~\cite{chung2017out} is utilized to measure lip synchronization quality, which is the most important indicator for talking head applications. To evaluate the realism and identity preservation of the results, we calculate the Frechet Inception Distance (FID)~\cite{heusel2017gans} and cosine similarity (CSIM) between the reference image and generated frames. Besides, the standard deviation of the generated head pose (both rotation and translation) is calculated to evaluate head pose diversity (HPD) better. 

\subsection{Implementation Details}
In our experiments, we use FaceVerse~\cite{wang2022faceverse}, the state-of-the-art single image reconstruction method to recover the video and obtain the ground truth blendshapes and meshes for supervision. During training, the Audio-To-Mesh stage and Mesh-To-Video stage are trained separately. Specifically, the BlendShape and Vertex Offset Generator and Learnable Head Pose Codebook in the Audio-To-Mesh stage are also trained separately. During inference, our model can work in an end-to-end manner by cascading the above two stages. For optimization, the Adam optimizer ~\cite{kinga2015method} is used with the learning rate $1\times10^{-4}$ and $1\times10^{-5}$ for two stage, respectively. And the total training costs 2 days on 8 NVIDIA V100 GPUs. More details about training and network architecture can be referred to in the supplementary material.

\subsection{Comparison with state-of-the-art methods}

\begin{table*}[]
\renewcommand\arraystretch{1.2}
\resizebox{\linewidth}{!}{

\begin{tabular}{l|c|cccc|ccc}
\hline
\multirow{2}{*}{Method} & \multirow{2}{*}{\begin{tabular}[c]{@{}c@{}}Head Pose\\ Generation\end{tabular}} & \multicolumn{4}{c|}{Same-Identity Reconstruction}                                         & \multicolumn{3}{c}{Cross-Identity Dubbing}                    \\ \cline{3-9} 
                        &                                                                                 & \multicolumn{1}{c}{SyncNet $\uparrow$} & \multicolumn{1}{c}{FID $\downarrow$} & \multicolumn{1}{c}{CSIM $\uparrow$} & HPD $\uparrow$ & \multicolumn{1}{c}{SyncNet $\uparrow$} & \multicolumn{1}{c}{CSIM $\uparrow$} & HPD $\uparrow$ \\ \hline

Real Video              & \usym{2717}     & 7.838        & 0.000      & 1.000     & 0.217     & \usym{2717}               & \usym{2717}             & \usym{2717}            \\
SadTalker \cite{zhang2023sadtalker}               & \usym{1F5F8}     & 5.711        & 28.35      & 0.862     & 0.305     & 5.416          & 0.849        & 0.337       \\
TalkLip \cite{wang2023seeing}                 & \usym{2717}         & 5.503        & 23.18      & 0.713     & \usym{2717}          & 5.295          & 0.686        & \usym{2717}            \\
MakeItTalk \cite{zhou2020makelttalk}              & \usym{1F5F8}      & 3.346        & 33.73      & 0.845     & 0.286     & 3.128          & 0.840        & 0.291       \\
Wav2Lip \cite{prajwal2020lip}                 & \usym{2717}            & \textbf{6.757}        & 21.80      & 0.816     & \usym{2717}          & \textbf{6.127}          & 0.807        & \usym{2717}            \\
PC-AVS \cite{zhou2021pose}                  & \usym{2717}            & 6.404        & 84.67      & 0.674     & \usym{2717}          & 5.538          & 0.613        & \usym{2717}            \\
Ours                    & \usym{1F5F8}       & 6.684        & \textbf{20.32}      & \textbf{0.916}     & \textbf{0.437}     & 6.018          & \textbf{0.907}        & \textbf{0.497}       \\ \bottomrule
\end{tabular}
}
\caption{The quantitative comparison with several state-of-the-art talking head generation works. Note that our proposed VividTalk outperforms previous works in video quality, identity preservation, and head pose diversity. 
}
\label{tab: quantitative_comparison}
\end{table*}

\begin{figure*}[th]
    \centering
    \includegraphics[width=1.0\linewidth]{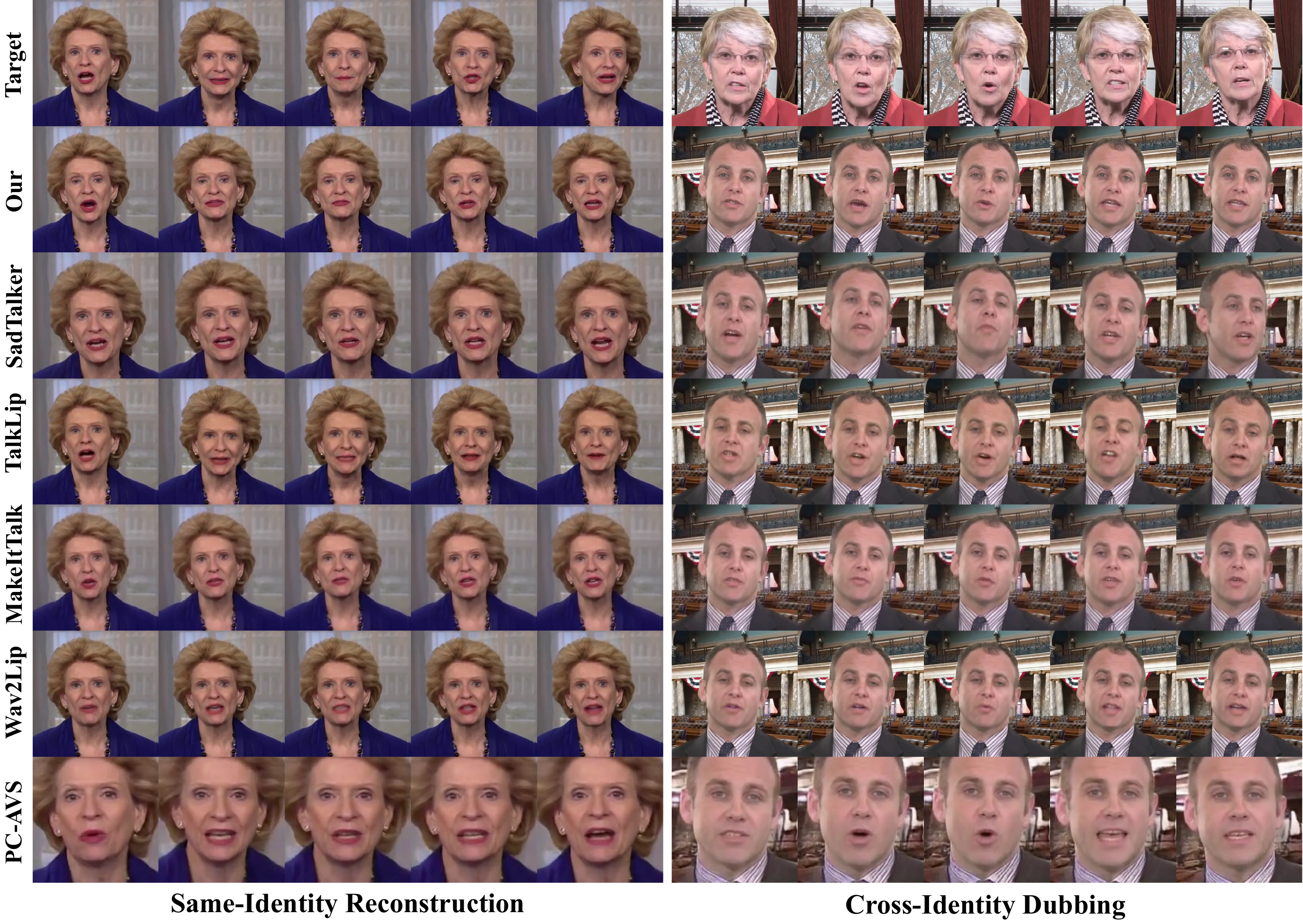}
    \vspace{-0.2in}
    \caption{The qualitative comparison results of our method and several state-of-the-art methods on talking head generation. SadTalker \cite{zhang2023sadtalker} and MakeItTalk \cite{zhou2020makelttalk} can generate results with a single image and audio as input. While TalkLip \cite{wang2023seeing}, Wav2Lip \cite{prajwal2020lip}, and PC-AVS \cite{zhou2021pose} need another video to provide the head poses for the final results. 
    }
    \vspace{-0.1in}
    \label{fig: qualitative_compare}
\end{figure*}

We qualitatively and quantitatively compare the proposed method to several prior state-of-the-art works on audio-driven talking head generation, including the SadTalker \cite{zhang2023sadtalker}, TalkLip \cite{wang2023seeing}, MakeItTalk \cite{zhou2020makelttalk}, Wav2Lip \cite{prajwal2020lip}, and PC-AVS \cite{zhou2021pose}. The experiments are conducted in Same-Identity Reconstruction and Cross-Identity Dubbing setting. In the Same-Identity Reconstruction setting, the audio signal and the reference image come from the same identity. While in the Cross-Identity Dubbing setting, videos non-existent in the world are generated because the audio comes from another person. 

\noindent \textbf{Qualitative Comparison. } Figure ~\ref{fig: qualitative_compare} demonstrates the visual results of our method and previous methods. It can be seen that SadTalker \cite{zhang2023sadtalker} fails to generate accurate fine-grained lip motion and is inferior to our video quality. This is because it only uses the blendshape as the intermediate representation which is insufficient to model the expressive facial motion. TalkLip \cite{wang2023seeing} generates blurry results and changes the skin color style to slightly yellow, which loses the identity information to a certain degree. MakeItTalk \cite{zhou2020makelttalk} can not generate accurate mouth shapes, especially in the Cross-Identity Dubbing setting. Wav2Lip \cite{prajwal2020lip} tends to synthesize blurry mouth regions, and output video with static head pose and eye movement when inputting a single reference image. PC-AVS \cite{zhou2021pose} requires a driven video as input and struggles for identity preservation. By contrast, our proposed method can generate high-quality talking head video with accurate lip-synchronized and expressive facial motion. 

\noindent \textbf{Quantitative Comparison. } As shown in Table ~\ref{tab: quantitative_comparison}, our method performs better in image quality and identity preservation, which is reflected by lower FID and higher CSIM metrics. Thanks to the novel learnable codebook mechanism, the head pose generated by our method is also more diverse and natural. Though the SyncNet score of our method is inferior to Wav2Lip \cite{prajwal2020lip}, our method can drive the reference image with single audio instead of video and generate frames in higher quality.

\subsection{User Studies}

\begin{table}[]
\renewcommand\arraystretch{1.2}
\resizebox{\linewidth}{!}{
\begin{tabular}{@{}l|c|c|c|c@{}}
\toprule
Method     & \begin{tabular}[c]{@{}c@{}}Lip\\ Sync\end{tabular} & \begin{tabular}[c]{@{}c@{}}Motion\\ Naturalness\end{tabular} & \begin{tabular}[c]{@{}c@{}}Identity\\ Preservation\end{tabular} & \begin{tabular}[c]{@{}c@{}}Overall\\ Quality\end{tabular} \\ \midrule
SadTalker \cite{zhang2023sadtalker}  &             3.891                                       &                 3.107                                             &                4.035                                                 &            3.626                                               \\
TalkLip \cite{wang2023seeing}  &             3.217                                       &                 \usym{2717}                                             &                3.891                                                 &            3.418                                               \\
MakeItTalk \cite{zhou2020makelttalk} &             2.836                                       &                 2.748                                             &                3.740                                                 &            2.914                                               \\
Wav2Lip \cite{prajwal2020lip}    &             2.751                                       &                 \usym{2717}                                             &                3.814                                                 &            2.471                                               \\
PC-AVS \cite{zhou2021pose}     &             3.106                                       &                 \usym{2717}                                             &                2.603                                                 &            2.513                                               \\
Ours       &             \textbf{4.315}                      &        \textbf{3.896}                                             &                        \textbf{4.618}                                         &           \textbf{4.307}                                                \\ \bottomrule
\end{tabular}
}
\caption{User study.}
\vspace{-0.1in}
\label{tab: user_study}
\end{table}

To further evaluate the proposed method, we conducted a user study with 20 volunteers to rate the videos generated by each method. For a fair comparison, 10 in-the-wild facial images with various characteristics and poses are selected as reference images, and 5 audio with diverse languages and speaking styles are chosen as driven signals, which are taken as the input of each method and generate 50 videos in total.  The volunteers are asked to rate each video between 1 and 5 (higher is better) in terms of lip synchronization, motion naturalness, identity preservation, and overall quality. As shown in Table~\ref{tab: user_study}, the final mean score of our method outperforms previous methods in all metrics, which indicates the superiority of our method. 

\subsection{Ablation Studies}
In this section, we conduct several ablation studies to verify the effectiveness of each design in proposed method.

\begin{figure}[th]
    \centering
    \includegraphics[width=1.0\linewidth]{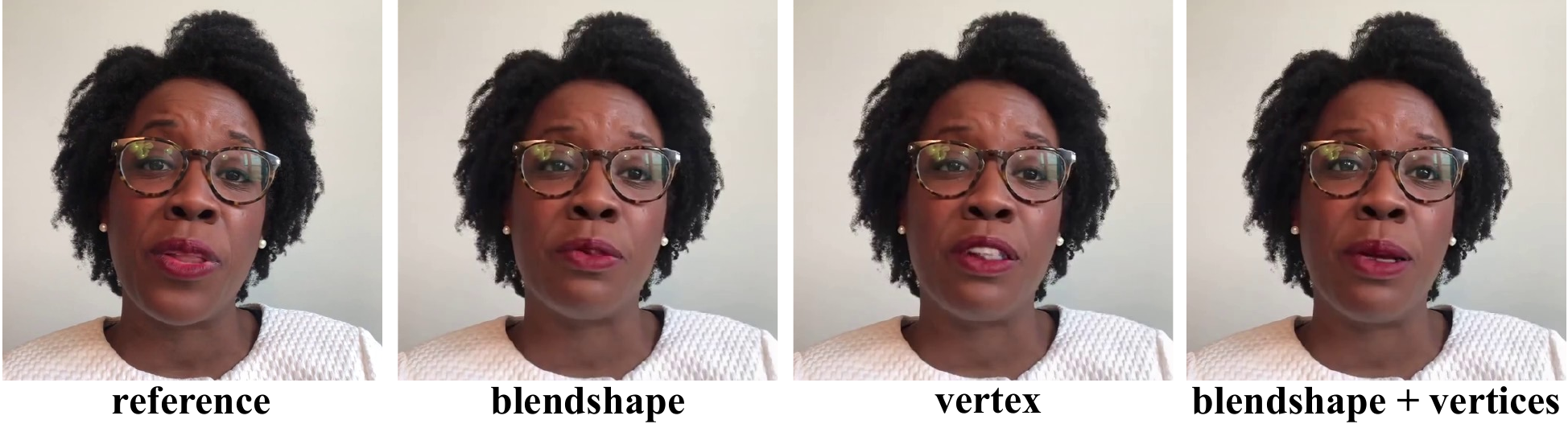}
    \vspace{-0.2in}
    \caption{Ablation about Intermediate representation.}
    \vspace{-0.1in}
    \label{fig: ablation_intermediate_representation}
\end{figure}

\noindent \textbf{Ablation about Intermediate Representation.} To verify the superiority of using both blendshape and vertex offset as the intermediate representation, we implement two variant models using either blendshape or vertex offset to generate 3D meshes from audio. The final driven results are shown in Figure~\ref{fig: ablation_intermediate_representation}. We can see that the method using blendshape only as the intermediate representation can model most facial expression motions well but not lip motion. The method using vertex offset as the intermediate representation can model the mouth shape better but lead to artifacts in the teeth region. By comparison, the method with both representation can generate accurate and fine-grained motions with high video quality maintained. 

\noindent \textbf{Ablation about Learnable Head Pose Codebook.} We also perform experiments to validate the design effectiveness of the Learnable Head Pose Codebook. On the one hand, we learn absolute instead of relative head pose from the audio. On the other hand, we remove the initial head pose $P^{0}$ as a condition in the mapping phase. As shown in Table~\ref{tab: abldtion_learnable_hpc}, learning absolute head pose leads to lower diversity, and our method without the initial head pose results in an unnatural visual effect. By contrast, our full method performs better in both evaluation metrics, indicating the benefits of our designs. 

\begin{table}[]
\renewcommand\arraystretch{1.2}
\resizebox{\linewidth}{!}{
\begin{tabular}{@{}l|c|c@{}}
\toprule
Method              & Diversity $\uparrow$ & Naturalness $\uparrow$ \\ \midrule
Absolute Head Pose Prediction &  0.379   &     3.641        \\
w/o Initial Head Pose    &  0.408   &     3.728        \\ 
Our Full        &  \textbf{0.437}   &    \textbf{3.896}        \\ 
\bottomrule
\end{tabular}
}
\caption{Ablation about Learnable Head Pose Codebook.}
\vspace{-0.1in}
\label{tab: abldtion_learnable_hpc}
\end{table}

\begin{figure}[th]
    \centering
    \includegraphics[width=1.0\linewidth]{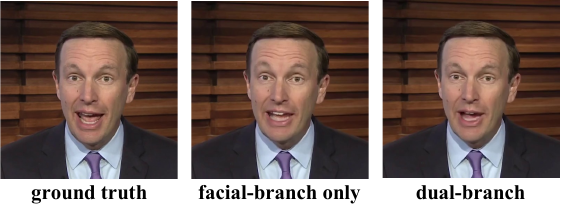}
    \vspace{-0.2in}
    \caption{Ablation about dual branch Motion-VAE.}
    \vspace{-0.1in}
    \label{fig: ablation_motion_vae}
\end{figure}

\noindent \textbf{Ablation about dual branch Motion-VAE.} We evaluate the proposed dual branch motion-vae in Mesh-To-Video stage regarding lip synchronization and video quality. Specifically, we designed a variant that keeps the facial motion branch only and removes the lip motion branch. As shown in Figure~\ref{fig: ablation_motion_vae}, the method without the lip-motion branch can not model the mouth shape accurately and generates frames with artifacts in the teeth area. In contrast,  the dual branch model can synthesize results well benefits from the enhancement of the lip-motion by lip-branch.

%% file: sec/5_con.tex
\section{Conclusion}
In this paper, we proposed VividTalk, a novel and generic framework supporting the generation of high-quality talking head videos with expressive facial expressions and natural head poses. For non-rigid expression motion, both blendshape and vertex are mapped as the intermediate representation to maximize the representation of the model, and an elaborate multi-branch generator is designed to model global and local facial motions individually. As for rigid head motion, a novel learnable head pose codebook with a two-phase training mechanism is proposed to synthesize natural results. Thanks to the dual branch motion-vae and generator, the driven meshes can be transformed into dense motion well and used to synthesize finale videos. Experiments demonstrate our method outperforms previous state-of-the-art methods and opens new avenues in many applications, such as digit human creation, video conferences, and so on. 